\documentclass{article}
\usepackage{spconf,amsmath,graphicx,cite}
\usepackage{hyperref}
\usepackage{tikz}
\usetikzlibrary{positioning}
\usetikzlibrary{calc}
\usetikzlibrary{arrows.meta}
\usepackage[tight]{subfigure}
\graphicspath{{../exp/adaptive_vc/test_clean_enroll/}{../exp/baseline/}{../exp/pwvc_s3/}}

\usepackage{xcolor,comment,xspace}
\newcommand{\marc}[1]{\textcolor{red}{Marc: #1}}

\newcommand{\const}{\emph{const}\xspace}
\newcommand{\rand}{\emph{random}\xspace}
\newcommand{\perm}{\emph{perm}\xspace}
\newcommand{\ignorant}{\emph{Ignorant}\xspace}
\newcommand{\semiIgnorant}{\emph{Semi-Informed}\xspace}
\newcommand{\informed}{\emph{Informed}\xspace}
\includecomment{altmarc}
\excludecomment{versiona}

\title{Evaluating voice conversion-based privacy protection against informed attackers} %

\name{\parbox{\linewidth}{\centering Brij Mohan Lal Srivastava$^{1}$, Nathalie Vauquier$^{1}$, Md Sahidullah$^{3}$, Aur\'elien Bellet$^{1}$, Marc Tommasi$^{2}$, Emmanuel Vincent$^{3}$\thanks{This work was supported in part by the European Union's Horizon 2020 Research and Innovation Program under Grant Agreement No. 825081 COMPRISE (\url{https://www.compriseh2020.eu/}) and by the French National Research Agency under project DEEP-PRIVACY (ANR-18-CE23-0018). Experiments presented in this paper were carried out using the Grid'5000 testbed, supported by a scientific interest group hosted by Inria and including CNRS, RENATER and several Universities as well as other organizations (see \url{https://www.grid5000.fr}).}}}
\address{$^1$INRIA, France\quad  $^2$Universit\'e de Lille, France \\ $^3$Universit\'e de Lorraine, CNRS, Inria, Loria, F-54000 Nancy, France}

\begin{document}
\ninept
\maketitle
\begin{abstract}

Speech data conveys sensitive speaker attributes like identity or accent. With a small amount of found data, such attributes can be inferred and exploited for malicious purposes: voice cloning, spoofing, etc. Anonymization aims to make the data {\it unlinkable}, i.e., ensure that no utterance can be linked to its original speaker. In this paper, we investigate anonymization methods based on voice conversion. In contrast to prior work, we argue that various {\it linkage} attacks can be designed depending on the attackers' knowledge about the anonymization scheme. We compare two frequency warping-based conversion methods and a deep learning based method in three attack scenarios. The utility of converted speech is measured via the word error rate achieved by automatic speech recognition, while privacy protection is assessed by the increase in equal error rate achieved by state-of-the-art i-vector or x-vector based speaker verification. Our results show that voice conversion schemes are unable to effectively protect against an attacker that has extensive knowledge of the type of conversion and how it has been applied, but may provide some protection against less knowledgeable attackers.

\end{abstract}
\section{Introduction}
\label{sec:intro}

Speech is a behavioural biometric characteristic of human beings~\cite{jain2000biometric}, which can produce distinguishing and repeatable biometric features. Dramatic improvements in speech synthesis~\cite{szekely2019spontaneous}, voice cloning~\cite{vestman2020voice,lorenzo2018can} and speaker recognition~\cite{snyder2018x} that leverage ``found data" pose severe privacy threats to the users of speech interfaces~\cite{kinnunen2017asvspoof}. According to the ISO/IEC International Standard 24745 on biometric information protection~\cite{iso24745}, biometric references must be {\it irreversible} and {\it unlinkable} for full privacy protection.
Anonymization or de-identification~
\cite{nautsch2019preserving,bahmaninezhad2018convolutional,srivastava2019privacy} refers to the task of concealing the speaker's identity while retaining the linguistic content, thereby making the data {\it unlinkable}~\cite{gomez2017general}. In this work, we consider the following threat model: given a public dataset of (supposedly) anonymized speech, an attacker records/finds a sample of speech of a target user and attempts to guess which utterances in the anonymized dataset are spoken by the target user.
A good anonymization scheme should prevent such \emph{linkage attacks}
from being successful,
while preserving the perceived speech naturalness and intelligibility and/or
the performance of downstream tasks such as automatic speech recognition (ASR).

Fang et al.~\cite{fang2019speaker} classify speaker anonymization methods into two categories: \textit{physical} vs.\ \textit{logical}. Physical methods perturb speech in the physical space by adding acoustic noise, while logical methods apply a transformation to the recorded signal. Among the latter, voice conversion (VC) methods have been traditionally exploited as a way to map the input voice (\emph{source}) into that of another speaker (\emph{target}) \cite{ribaric2016identification,pobar2014online,qian2018hidebehind}. In contrast to feature-domain approaches \cite{srivastava2019privacy}, the output of VC remains a speech waveform and it may be used for listening or transcription purposes. The anonymized speech should thus sound as natural and intelligible as possible~\cite{jin2009speaker}.

Crucially, all past studies assumed a weak attack scenario where the attacker is unaware that an anonymization method has been applied to the found data~\cite{fang2019speaker}. This raises the concern that the privacy protection may entirely rely on the secrecy of the design and implementation of the anonymization scheme, a principle known as ``security by obscurity''~\cite{mercuri2003security} that has long been rejected by the security community.
There is therefore a strong need to evaluate the robustness of the anonymization to the knowledge that the adversary may have about the transformation. In practice, such knowledge may for instance be acquired by inspecting the code embedded in the user's device or in an open-source implementation.

As opposed to past studies, we consider different linkage attacks
depending on the attacker's knowledge of the anonymization method. At one end
of the spectrum, an \ignorant attacker is unaware of the speech transformation
being applied, while at the other end an \informed attacker can leverage complete knowledge of the transformation algorithm. A \semiIgnorant attacker may know the voice transformation algorithm but not its parameter values.
In our experiments, we evaluate three VC methods with different target speaker selection strategies in various attack scenarios to study unlinkability in the spirit of ISO/IEC 30136 standard~\cite{iso30136}.
In each scenario, we assess how well each method protects the speaker identity
against attackers that leverage state-of-the-art speaker verification
techniques based on i-vectors~\cite{dehak2010front} or x-vectors~
\cite{snyder2018x} to design linkage attacks.
We also report the \textit{word error rate} (WER) achieved by a state-of-the-art end-to-end automatic speech recognizer~\cite{watanabe2018espnet}. While a formal listening test is beyond the scope of this paper, we make a few samples of converted speech available for informal comparison
.\footnote{\url{https://github.com/brijmohan/adaptive_voice_conversion/tree/master/samples}}

In Section \ref{sec:vc-methods}, we describe the three VC methods we evaluate in the context of anonymization. Section \ref{sec:attacker} introduces the target speaker selection strategies and the attack scenarios. Section \ref{sec:exp} presents the experimental settings and the results. We conclude in Section \ref{sec:conc}.


\section{Voice conversion methods}
\label{sec:vc-methods}

The criteria for selecting the VC methods in our study are that they must be {\bf 1) non-parallel}, i.e., do not require a parallel corpus of sentences uttered by both the source and target speakers for training --- this is important 
from a privacy perspective since there exist few parallel corpora and selecting openly available targets would increase the risk of an inversion attack; {\bf 2) many-to-many}, i.e., allow conversion between arbitrary sources and targets so that any speaker in a large corpus can be selected as the target
; {\bf 3) source- and language-independent}, i.e., do not require enrollment sentences for the source speaker and do not rely on language-specific ASR or phoneme classification --- this is important from a usability perspective as it frees the user from the burden of enrolling and it is applicable to any language (including under-resourced ones), and from a privacy perspective since enrollment translates into the storage of a voiceprint which poses even greater privacy threats.

The third criterion is quite strict: many VC methods, such as StarGAN-VC~\cite{kameoka2018stargan} or the ASR-based method in \cite{fang2019speaker}, do not satisfy it. 
We found that the vocal tract length normalization (VTLN) based methods in \cite{qian2018hidebehind,sundermann2003vtln} and the one-shot method in \cite{chou2019one} satisfy all criteria. 
In this paper, we use models trained over English speech~\cite{panayotov2015librispeech} but do not use any other linguistic resources such as transcriptions.


\subsection{VoiceMask}
VoiceMask is described in~\cite{qian2018hidebehind} as the frequency warping method based on the composition of a log-bilinear function, expressed as $f(\omega, \alpha) = |-i\ln\frac{z - \alpha}{1 - \alpha z}|$, and a quadratic function, given by $g(\omega, \beta) = \omega + \beta (\frac{\omega}{\pi} - (\frac{\omega}{\pi})^2)$. Here $\omega \in [0, \pi]$ is the normalized frequency, $\alpha \in [-1, 1]$ is the warping factor for the bilinear function, $z = {\rm e}^{i\omega}$, and $\beta > 0$ is the warping factor for the quadratic function. Therefore, the warping function is of the form $g(f(\omega,\alpha),\beta)$. The two parameters, $\alpha$ and $\beta$, are chosen uniformly at random from a predefined range which is found to produce intelligible speech while perceptually concealing the speaker identity. In the following, we apply this transform to the spectral envelope rather than the pitch-synchronous spectrum as in the original paper. In addition, we apply logarithm Gaussian normalized pitch transformation (see~\cite{liu2007high}) so as to match the pitch statistics of a target speaker\footnote{Strictly speaking, VoiceMask is a voice transformation method rather than a VC method: pitch is converted from the source speaker to a target speaker, but the spectral envelope is not related to a particular target speaker.}. 

The authors claim that this transformation is difficult to inverse when the parameter values are unknown because they are randomly selected from a large interval. However, VoiceMask uses the same parameter values to warp the spectra at each time step of the utterance. This approach is quite limited to conceal the identity of the source speaker and to mimic the target speaker because it warps the entire frequency axis in a single direction.

\subsection{VTLN-based voice conversion}
VTLN-based VC \cite{sundermann2003vtln} represents each speaker by a set of centroid spectra extracted using the CheapTrick~\cite{MORISE20151} algorithm for $k$ pseudo-phonetic classes. These classes are learned in an unsupervised fashion by clustering all speech frames of all utterances from this speaker. For each class of the source speaker, the procedure finds the class of the target speaker and the warping parameters that minimize the distance between the transformed source centroid spectrum and the target centroid spectrum. All speech frames in that class are then warped using a power function. Similarly to above, we apply this warping to the spectral envelope and also perform Gaussian normalized pitch transformation so as to match the pitch statistics of the target. Compared to VoiceMask, this approach warps the frequency axis in different directions over time.
The parameters of this method include the number of classes $k$ and the chosen target speaker.

\subsection{Disentangled representation based voice conversion}
The third approach is based on disentangled representation of speech as proposed in~\cite{chou2019one,ulyanov2017improved}. The core idea is that speaker information is statically present throughout the utterance but content information is dynamic. This approach is based on a neural network transformation and uses a {\it speaker encoder} and a {\it content encoder} to separate the factors of variation corresponding to speaker and content information. The only parameter of this method is the chosen target speaker.


\begin{versiona}
  
\marc{Should be explicitly said that it is a neural network based transformation. I don't understand the use of channel here. Is it useful? }
The third approach is based on disentangled representations of speech as proposed in~\cite{chou2019one}. The central idea of this work is that speaker/channel information is statically present throughout the utterance but content information is dynamic. Hence, as shown in~\cite{ulyanov2017improved}, instance-wide static information such as image contrast can be removed by using instance normalization layer in the network. Similar idea can be applied to remove the speaker/channel information and just keep the content information.




This approach uses a {\it speaker encoder} and a {\it content encoder} to separate the factors of variation corresponding to speaker and content information. The content encoder uses instance normalization to cancel out the speaker/channel effect. \marc{We announce that it doesn't work so why do we use this transformation? So either we do not use that transform or we remove the sentence (or maybe moderate it) and the plots?} Although with a simple TSNE plot we found that the embeddings obtained from content encoder still possess considerable amount of speaker information, which we refer to as incomplete disentanglement.

Figure~\ref{fig:dar_enroll_tsne} shows TSNE plot for speaker and content embeddings obtained for five male and five female speakers present in the enrollment set. The model used here was trained over LibriTTS~\cite{zen2019libritts} corpus as described in section~\ref{sec:exp}.

\begin{figure}[h!]
\centering
	\subfigure[Speaker embeddings]{\label{fig:spk_enroll_dar}\includegraphics[width=0.45\linewidth]{disentangled_spks_embedding.png}}
	\subfigure[Content embeddings]{\label{fig:content_enroll_dar}\includegraphics[width=0.45\linewidth]{disentangled_content_embedding.png}}
  \caption{TSNE plot for one-shot speaker and content embeddings obtained for unseen data. Same Color indicates same speaker. Dots are male, triangles are female. Speaker embeddings are well-clustered. Content embeddings still contain neighbourhood as same speakers.}
  \label{fig:dar_enroll_tsne}
  \vspace{-2em}
\end{figure}
\end{versiona}


\section{Target selection strategies and Attackers}
\label{sec:attacker}


\begin{versiona}
Let $f$ be defined as the VC function parametrized by $\theta$. Then $f(i, j;\theta)$ maps an utterance of speaker $i$ to speaker $j$. The target speaker selection strategy is given by $\sigma(i)=j$, which eventually generates a mapping $i \rightarrow j$. These are the key ingredients of a ``privacy-by-design'' algorithm. We define some specific strategies used in this paper.
\end{versiona}
\begin{altmarc}
In this study, we consider that the VC function and the sets of possible
parameter values are known to all users. Each user records his/her voice on
his/her device and applies a VC scheme locally before sending it to a public
database. In the threat model we consider, an attacker then 
performs a linkage attack to try to identify which converted utterances
in this public database are spoken by a particular user. To this end,
we assume that the attacker
has access to a small amount of found speech from this user (and potentially some additional public resources, such as benchmark speech processing datasets to train generic speaker models).

In the following, we define three parameter selection (a.k.a. target selection) strategies for the three VC methods above, which can be seen as key ingredients of a ``private-by-design'' speech processing system. We then describe the knowledge that an attacker trying to compromise the system could have about the VC function and the target selection strategy. 
\end{altmarc}

\subsection{Target selection strategies}
\label{sec:ano_strats}

\begin{altmarc}

  We consider three possible target selection strategies. In strategy \const, the VC function is constant across all users and all utterances. This means choosing a unique target speaker and, in the case of VoiceMask, fixed values for $\alpha$ and $\beta$. In strategy \perm, the conversion parameters are chosen at random once by each user. In other words, when a user downloads the VC module on his/her device, he/she selects a personal target speaker and, in the case of VoiceMask, personal random values for $\alpha$ and $\beta$. Finally, in the \rand strategy, each time a user applies VC to an utterance, a random set of parameters is drawn, i.e., a random target speaker is selected and, in the case of VoiceMask, random values are drawn for $\alpha$ and $\beta$.
  
\end{altmarc}

\begin{versiona}
In strategy \const, $\sigma(.)$ is a constant function, i.e. $j$ is always the same. In strategy \perm, for every speaker $i$ we pick a speaker $\sigma(i)=j$ among $k=100$ target speakers. In strategy \rand, for every utterance $u_i$ belonging to speaker $i$, we pick a speaker $\sigma(u_i)=j$ among $k=100$ target speakers. Note that $\theta$ is either learned from data (as in case of GMM-based VTLN and one-shot approach) or derived from pre-existing knowledge (Voicemask). Function $\sigma(.)$ must be selected carefully based on its {\it privacy-preserving properties}. These properties enable function $f(.)$ to successfully anonymize the speaker's identity and increase the error of speaker verification network.

Note that strategy \rand implies more randomness, which could result in increased privacy. However some averaging-based attacks could take benefit of this kind of VC. Therefore, in some cases we have kept the three strategies.

\end{versiona}




\subsection{Attackers' knowledge}
\label{sec:knowledge}

\begin{altmarc}
We define the types of attackers based on the extent of their knowledge about the VC function and its parameters.     
An \ignorant attacker is not aware that VC has been applied at all. In contrast, an \informed attacker knows the VC method and its exact parameter values (i.e., the chosen target speaker and the values of $\alpha$ and $\beta$). One may argue that an \informed attacker is not very realistic (except for the \const strategy), while an \ignorant attacker is very weak. Between these two extreme cases, various types of attackers can be defined. For instance, we consider a \semiIgnorant attacker who knows the chosen VC method (VoiceMask, VTLN, or disentangled representation) and the target selection strategy (\const, \perm, or \rand), but not the actual target (i.e., the actual target speaker or the value of $\alpha$ and $\beta$). This is arguably more realistic since the VC algorithm and the target selection strategy may be open-source, while (except for the \const strategy) the target chosen by the user is much less easily accessible.

It is important to note that many concrete instances of attackers of the above types can be designed, and finding out the ``best'' attacker of a particular type is a hard problem.
In the experiments section, we propose attackers exploiting these different levels of knowledge based on the assumptions defined above. A more exhaustive investigation of the design of attackers is left for future work.

\end{altmarc}
\begin{versiona}
We define the attackers based on the extent of their knowledge about function $f(.)$ and $\sigma(.)$. The strongest attacker knows the VC function $f$, its parameters $\theta$ and target selection strategy $\sigma$. It leverages this knowledge during training and verification of speaker identification models. We refer to this attacker in scenario 1. The weakest attacker does not know anything about these functions and processes the incoming speech as it would process the untransformed speech. We refer to this attacker in scenario 2. There can be several other types of attackers based on their partial knowledge about function $f$, parameter $\theta$ and mapping function $\sigma$, but in this study, we only compare the strongest and the weakest attacker.
\end{versiona}


\begin{versiona}
We roughly define the four types of attackers:
\begin{enumerate}
\item \textbf{Weak}: The attacker knows nothing. As a specific case, they use untranformed corpus to train the x-vector/i-vector model and to enroll the speakers, while speaker verification trials are conducted with the converted data.
\item \textbf{Moderate weak}: The attacker only knows $f$. They have to guess/choose some $\theta$, $\sigma$ and mapping $j$. We do not study this case in this paper due to broad range of experiments needed to simulate it.
\item \textbf{Moderate strong}: The attacker knows $f$, $\theta$ and also $\sigma$, but does not know the exact mapping between $i$ and $j$. So for instance in Strategy 2, speaker $i$ will be not necessarily mapped to the same user for enrollment than in the trial set. In this case, the attacker uses the transformed corpus to train the x-vector/i-vector model and to enroll the speakers. But the trial corpus as well as to conduct speaker verification trials. They will choose a $j$ (at random).
\item \textbf{Strong}: The attacker knows $f$, $\theta$ and $\sigma$. So for instance in Strategy 2, speaker $i$ will be mapped to the same user $j$ for enrollment  than in the trial set. The attacker uses the transformed corpus to train the x-vector/i-vector model, to enroll the speakers as well as to conduct speaker verification trials.
\end{enumerate}
Note that we define the attacker knowledge but simulating specific speaker is a hard problem due to vast number of potential design choices. Here we choose to simulate the attacker based on the assumptions defined above. We defer to exhaustively investigate the attackers as part of future work.
\end{versiona}

\begin{table*}[t!]
  \renewcommand{\arraystretch}{1.2}
  \centering
  \caption{EER (\%) achieved using x-vector based speaker verification.}
  \footnotesize
  \begin{tabular}{|l|c|c|c|c|c|c|c|}\hline
    &\textbf{VoiceMask}&\multicolumn{3}{c|}{\textbf{VTLN-based VC}}&\multicolumn{3}{c|}{\textbf{Disentangl.-based VC}}\\ \cline{2-8}
    \textbf{Attackers~$\downarrow$ / Strategies~$\rightarrow$} & \rand & \const& \perm& \rand & \const & \perm & \rand\\  \hline
    \informed & 5.01 & 4.71  & 3.91  & 6.32  & 4.71  & 0.20 & 5.52  \\ \hline
    \semiIgnorant & - & 12.84  & 23.37  & 6.32  & 13.64  & 43.03 & 5.42  \\ \hline
    \ignorant & 28.69 & 24.27 & 30.99 & 27.38 & 27.68 & 32.20 & 30.59 \\ \hline
  \end{tabular}
  \footnotesize
  \label{tab:xvector_results}
  \vspace{-1.25em}
\end{table*}

\begin{table*}[t!]
  \renewcommand{\arraystretch}{1.2}
  \centering
  \caption{EER (\%) achieved using i-vector based speaker verification.}
  \footnotesize
  \begin{tabular}{|l|c|c|c|c|c|c|c|}\hline
    &\textbf{VoiceMask}&\multicolumn{3}{c|}{\textbf{VTLN-based VC}}&\multicolumn{3}{c|}{\textbf{Disentangl.-based VC}}\\ \cline{2-8}
    \textbf{Attackers~$\downarrow$ / Strategies~$\rightarrow$} & \rand & \const& \perm& \rand & \const & \perm & \rand\\  \hline
    \informed & 8.22 & 6.22  & 10.23  & 9.84  & 4.71  & 0.20 & 11.03  \\ \hline
    \semiIgnorant & - & 18.25  & 31.49  & 18.76  & 15.65  & 43.93 & 10.53  \\ \hline
    \ignorant & 50.55 & 26.08 & 49.15 & 49.15 & 49.95 & 47.74 & 49.85 \\ \hline
  \end{tabular}
  \footnotesize
  \label{tab:ivector_results}
  \vspace{-1.25em}
\end{table*}

\begin{table*}[t!]
  \centering
  \caption{WER (\%) achieved using end-to-end ASR.}
  \footnotesize
  \begin{tabular}{|l|c|c|c|c|c|c|c|}\hline
    &\textbf{VoiceMask}&\multicolumn{3}{c|}{\textbf{VTLN-based VC}}&\multicolumn{3}{c|}{\textbf{Disentangl.-based VC}}\\ \cline{2-8}
    \textbf{Subset~$\downarrow$ / Strategies~$\rightarrow$} & \rand & \const& \perm& \rand & \const & \perm & \rand\\  \hline
    test-clean   & 18.1 & 19.8  & 18.4  & 15.9  & 41.5  & 23.7 & 115.1 \\ \hline
  \end{tabular}
  \footnotesize
  \label{tab:wer}
  \vspace{-1.25em}
\end{table*}

\section{Experiments and Results}
\label{sec:exp}



\subsection{Data and evaluation setup}
\label{sec:eval}

All experiments are performed on the LibriSpeech corpus~\cite{panayotov2015librispeech}. We use the 460~h clean training set (\emph{train-clean-100} + \emph{train-clean-360}), which contains 1,172 speakers, to train the disentanglement transform. Out of the \emph{test-clean} set, we create an \emph{enrollment} set (438 utterances) and a \emph{trial} set (1,496 utterances) with different utterances from the same 29 speakers (13 male and 16 female, not in the training set) considered as source speakers. The target speakers for all three VC methods are randomly picked from the training and \emph{test-clean} sets. See \cite{srivastava2019privacy} for more details. 

For each VC method and target selection strategy, all utterances in the trial
set are mapped to possibly different target speakers in the training or trial
set. The converted trial set serves as the public database that attackers want
to de-anonymize by designing a linkage attack. To this end, attackers
have access to the enrollment set which serves as the found data used to model the speakers in the trial set.

The attackers also have access to the 460~h training set to train state-of-the-art speaker verification methods based on x-vectors~\cite{snyder2018x} and i-vectors, which are stronger than the Gaussian mixture model-universal background model (GMM-UBM) based method used in the seminal work of~\cite{jin2009speaker}. We adapt the {\it sre16} Kaldi recipe for training x-vectors and i-vectors to LibriSpeech\footnote{\url{https://github.com/brijmohan/kaldi/tree/master/egs/librispeech\_spkv/v2}}.
We use a smaller network architecture for x-vector computation than the original recipe. Specifically, compared to the architecture in \cite[Table~1]{snyder2018x}, we remove the {\it frame4}, {\it frame5} and {\it segment7} layers, thereby also reducing the {\it stats pooling} layer to 512$T$$\times$1024 and the {\it segment6} layer to 1024$\times$512. Here $T$ refers to the utterance-level context. This reduced architecture performs slightly better on LibriSpeech than the architecture in the original recipe.
We give more details on the different attackers in Section~\ref{sec:attackers}.

Finally, we evaluate the utility of each VC method in terms of the resulting ASR performance on the converted data. We use a hybrid connectionist temporal classification (CTC) and attention based encoder-decoder~\cite{watanabe2018espnet} trained on the converted 460~h training set using the standard recipe for LibriSpeech provided in ESPnet\footnote{\url{https://espnet.github.io/espnet/}}.
\subsection{Voice conversion settings}
\label{sec:exp_setting}






\textbf{VoiceMask.}
Pitch, aperiodicity and spectral envelope are extracted using the pyworld vocoder\footnote{\url{https://github.com/JeremyCCHsu/Python-Wrapper-for-World-Vocoder}}.
We follow strategy \rand only. We sample $\alpha$ uniformly such that $ |\alpha| \in [0.08,0.10]$ then $\beta$ in $[-2,2]$ such that $ 0.32 \leq dist_{f_{\alpha, \beta}} \leq 0.40 $ where $dist_{f_{\alpha, \beta}} = \int_0^\pi |f_{\alpha, \beta}(\omega) - \omega| $ is the distortion strength of the warping function. 
These ranges are provided by VoiceMask's authors in~\cite{qian2018hidebehind} since they produce most intelligible output. A subset of 100 target speakers is randomly selected and, for every utterance, pitch is transformed so as to match a random speaker within that subset. Other target selections strategies have not been applied because fixed values for $\alpha$ and $\beta$ (whether speaker-dependent or not) are prone to inversion attacks. 

\textbf{VTLN-based VC.}
Pitch, aperiodicity and spectral envelope are extracted using the pyworld vocoder. For each speaker, we collect speech frames using energy-based voice activity detection (VAD) with a threshold of 0.06, and we cluster their spectral envelopes via k-means with $k = 8$. 
In strategy \const, only one target speaker is selected. In \perm, we draw a random subset of 100 target speakers and, for each source speaker, we select a random target within it. In \rand, we draw a random subset of 100 target speakers and, for each source utterance, we select a random target within it.

\textbf{Disentangled representation based VC.}
We use a publicly available implementation of this method\footnote{\url{https://github.com/jjery2243542/adaptive\_voice\_conversion}}.
As per the authors' suggestion in the preprocessing script, we train the disentanglement models (speaker encoder, content encoder, decoder) over
the \emph{train-clean-100} subset of the LibriTTS corpus (itself a subset of the 460~h training set of LibriSpeech), with a batch size of 128 and learning rate of 0.0005 for 500,000 iterations. 
All three target selection strategies are applied similarly to VTLN-based VC except that only the source utterance and one random utterance from the target speaker are used as inputs to the content and speaker encoders, respectively. Other utterances from the source and targets speakers are unused.


\subsection{Attackers}
\label{sec:attackers}

We have implemented several attackers depending on the choice of the VC algorithm and the target selection strategy as well as the extent of the attacker's knowledge (\informed, \semiIgnorant or \ignorant).
Our \ignorant attacker is unaware of the VC step: he/she simply trains x-vector/i-vector models on the untransformed training set, and applies them to the untransformed enrollment set.
Our \semiIgnorant attacker knows the VC algorithm and the target selection strategy (\const, \rand or \perm) but not the particular choices of targets. He/she applies this strategy to the training and enrollment sets by drawing random target speakers from the subset of 100 target speakers used by the VC method (we assume that the value of $k$ in VTLN is known to the attacker). As a result, the training and enrollment data are converted in a similar way as the trial data, but the target speaker associated with every speaker in the enrollment set is typically different from that associated with the same speaker in the converted trial set.
Finally, our \informed attacker has access to the actual VC models and target choices used to anonymize the trial set, so it converts the training and enrollment sets accordingly.

In our preliminary experiments, we also considered attackers who convert the enrollment set only and use x-vector/i-vector models trained on the untransformed training set. Unsurprisingly, we found that this leads to significantly larger {\it equal error rates} (EER) than re-training the x-vector/i-vector model (which can easily be done by the attacker using public benchmark data). Therefore, we do not report results for such attackers below.


\subsection{Results and discussion}
\label{sec:results}

We first train and apply the ASR and speaker verification systems on the original (untransformed) data for baseline performance. We obtain an EER of $4.61\%$ and $4.31\%$ for i-vector and x-vector, respectively, and a WER of $9.4\%$ for ASR.

Tables~\ref{tab:xvector_results} and~\ref{tab:ivector_results} present the EER for x-vector and i-vector based speaker verification for the three attackers and the various VC methods and target selection strategies. Interestingly, the \informed attacker achieves similar or even slightly lower EER than the baseline. This indicates that, when the attacker has complete knowledge of the VC scheme and target speaker mapping, none of the VC methods is able to protect the speaker identity.
While an attacker with such complete knowledge is not very realistic in most practical cases, our results show that speaker information has not been totally removed and is somehow still present in the converted speech.

For the more realistic \semiIgnorant attacker,
we observe that strategy \perm is quite effective in protecting privacy and shows the highest gains in EER. This is due to the fact that the target speaker in the enrolled data may not be same as the one in trial, hence greater confusion is induced during inference. We also notice that strategy \rand is not much affected by the change of speaker mapping, which is intuitive because in this case the utterances are already being mapped randomly to different speakers. Such mapping would be ineffective due to averaging of randomness. Strategy \const is also slightly affected by the change of mapping because the training and enrollment speaker is not same as that of test speaker, but the effect is not as significant as strategy \perm.

\begin{figure}[h!]
\centering
  \includegraphics[width=\linewidth]{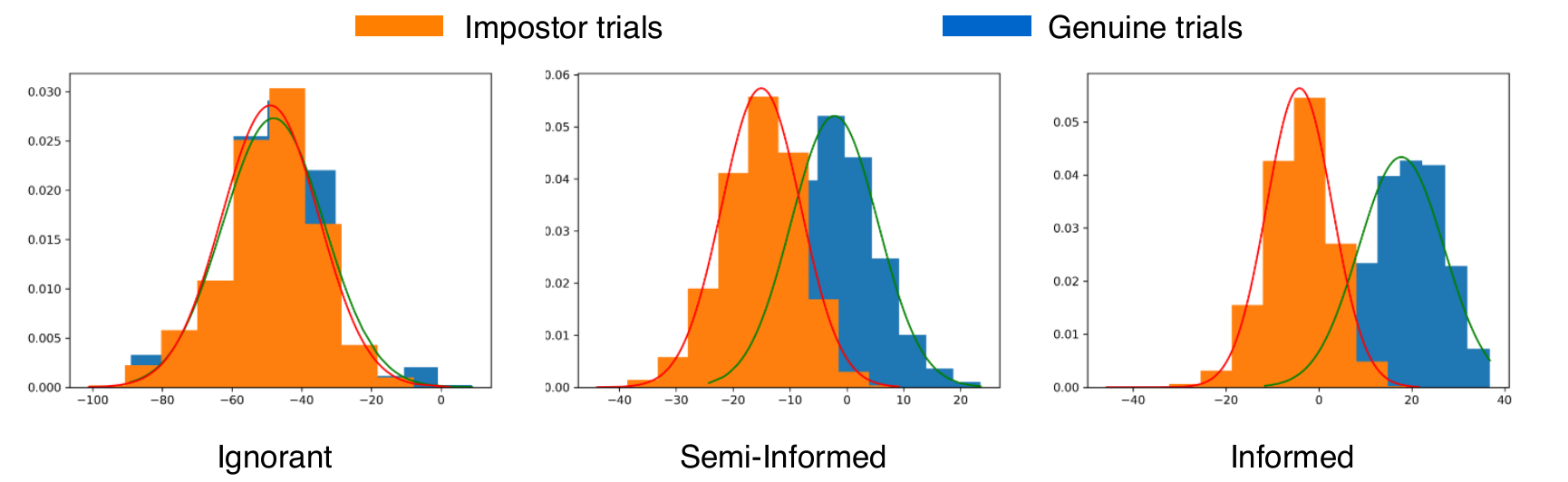}
  \caption{I-vector score distribution for trials conducted on VTLN (strategy \rand) converted data by \ignorant, \semiIgnorant, or \informed attackers. The orange distribution indicates impostor scores, while the blue distribution indicates genuine scores. The crossing between the two curves indicates the threshold for EER. More overlap means greater confusion, hence greater privacy protection.}
  \label{fig:score_dist}
\end{figure}

Consistently with past results in the literature, the \ignorant attacker
performs worst in terms of EER. This confirms that, when the attacker is oblivious to the privacy-preserving mechanism, we can protect speaker identity completely. Figure~\ref{fig:score_dist} shows the distribution of i-vector PLDA scores for genuine and impostor trials, i.e., the log-likelihood ratios between {\it same-speaker} and {\it different-speaker} hypotheses. For full unlinkability, the distributions of genuine and impostor scores must be identical. We observe that the overlap between the two distributions decreases as we move from the \ignorant to the \informed attacker, hence increasing linkability.


Table~\ref{tab:wer} gives the WER obtained for each VC method, which we
use as a proxy for the usefulness of the converted speech. Note that there is
no difference between converted data in different attack scenarios, hence the WER does not depend on the attacker.
VoiceMask and VTLN-based VC achieve reasonable WER compared to the untransformed data, while the disentangled representation based VC produces unreasonably high WER.
Note that these WERs are achieved when ASR is trained solely using converted data. In practice, many techniques can be used optimize the WER, such as using converted data to augment clean data.
\section{Conclusion and future work}
\label{sec:conc}

We investigated the use of VC methods to protect the
privacy of speakers by concealing their identity. We formally
defined target speaker selection strategies and linkage attack scenarios based on the knowledge of attacker.
Our experimental results indicate that both aspects play an important role in
the strength of the protection. Simple methods such as
VTLN-based VC with appropriate target selection
strategy can provide reasonable protection against linkage attacks with
partial
knowledge.

Our characterization of strategies and attack scenarios opens up several avenues for future research.
To increase the naturalness of converted speech, we can explore intra-gender
VC as well as the use of a supervised phonetic
classifier in VTLN.
We also plan to conduct experiments with a broader range of attackers and use standard local and global unlinkability metrics~\cite{gomez2017general} to
precisely evaluate the privacy protection in various scenarios.
More generally, designing a
privacy-preserving transformation which induces a large overlap between genuine
and impostor distributions even in the \informed attack scenario remains an open question. In the case
of disentangled representations, this calls for avoiding  any
leakage of private attributes into the content embeddings.



\vfill\pagebreak

\bibliographystyle{IEEEbib}
\bibliography{strings,refs}

\end{document}